\documentclass[a4paper,twoside]{article}

\usepackage{epsfig}
\usepackage{subcaption}
\usepackage{calc}
\usepackage{amssymb}
\usepackage{amstext}
\usepackage{amsmath}
\usepackage{amsthm}
\usepackage{multicol}
\usepackage{pslatex}
\usepackage{apalike}
\usepackage{algorithm2e}
\usepackage{hyperref}  
\usepackage[bottom]{footmisc}
\usepackage{lineno}

\usepackage{graphicx}
\usepackage{booktabs}
\usepackage{pgfplots}
\pgfplotsset{compat=1.18}
\usepackage{color}
\usepackage{multirow}
\usepackage{tabularx}
\usepackage{tikz}
\usepackage[T1]{fontenc}
\usepackage{siunitx}
\usepackage[table]{xcolor}
\usepackage{placeins}
\usepackage{float}
\usepackage{makecell}
\usepackage[none]{hyphenat}
\usepackage{algorithm2e}  

\usepackage{SCITEPRESS}

\graphicspath{{./}{Chapters/}{Chapters/01_Introduction/Figures/}{Chapters/03_Method/Figures/}{Chapters/05_Results/Tables/}}

\makeatletter
\def\thickhline{%
\noalign{\ifnum0=`}\fi\hrule \@height \thickarrayrulewidth \futurelet \reserved@a\@xthickhline}
\def\@xthickhline{\ifx\reserved@a\thickhline \vskip\doublerulesep \vskip-\thickarrayrulewidth \fi \ifnum0=`{\fi}}
\makeatother

\newlength{\thickarrayrulewidth}
\definecolor{tu0}{rgb}{0.7451, 0.1176, 0.2353}
\definecolor{tu1}{rgb}{1.0000, 0.8039, 0.0000}
\definecolor{tu11}{rgb}{1.0000, 0.8627, 0.3020}
\definecolor{tu12}{rgb}{1.0000, 0.9020, 0.4980}
\definecolor{tu13}{rgb}{1.0000, 0.9412, 0.6980}
\definecolor{tu14}{rgb}{1.0000, 0.9608, 0.8000}
\definecolor{tu2}{rgb}{0.9804, 0.4314, 0.0000}
\definecolor{tu21}{rgb}{0.9882, 0.6039, 0.3020}
\definecolor{tu22}{rgb}{0.9882, 0.7137, 0.4980}
\definecolor{tu23}{rgb}{0.9922, 0.8275, 0.6980}
\definecolor{tu24}{rgb}{0.9961, 0.8863, 0.8000}
\definecolor{tu3}{rgb}{0.6902, 0.0000, 0.2745}
\definecolor{tu31}{rgb}{0.7529, 0.2000, 0.4196}
\definecolor{tu32}{rgb}{0.8431, 0.4980, 0.6353}
\definecolor{tu33}{rgb}{0.9216, 0.7490, 0.8196}
\definecolor{tu34}{rgb}{0.9529, 0.8510, 0.8902}
\definecolor{tu4}{rgb}{0.4863, 0.8039, 0.9020}
\definecolor{tu41}{rgb}{0.6431, 0.8627, 0.9333}
\definecolor{tu42}{rgb}{0.7412, 0.9020, 0.9490}
\definecolor{tu43}{rgb}{0.8431, 0.9412, 0.9686}
\definecolor{tu44}{rgb}{0.8980, 0.9608, 0.9804}
\definecolor{tu5}{rgb}{0.0000, 0.5020, 0.7059}
\definecolor{tu51}{rgb}{0.3020, 0.6510, 0.7961}
\definecolor{tu52}{rgb}{0.5490, 0.7765, 0.8667}
\definecolor{tu53}{rgb}{0.7490, 0.8745, 0.9255}
\definecolor{tu54}{rgb}{0.8510, 0.9255, 0.9569}
\definecolor{tu6}{rgb}{0.0000, 0.3255, 0.4549}
\definecolor{tu61}{rgb}{0.2510, 0.4941, 0.5922}
\definecolor{tu62}{rgb}{0.5490, 0.6941, 0.7529}
\definecolor{tu63}{rgb}{0.7490, 0.8314, 0.8627}
\definecolor{tu64}{rgb}{0.8510, 0.8980, 0.9176}
\definecolor{tu7}{rgb}{0.7765, 0.9333, 0.0000}
\definecolor{tu71}{rgb}{0.8431, 0.9529, 0.3020}
\definecolor{tu72}{rgb}{0.8863, 0.9647, 0.4980}
\definecolor{tu73}{rgb}{0.9333, 0.9804, 0.6980}
\definecolor{tu74}{rgb}{0.9569, 0.9882, 0.8000}
\definecolor{tu8}{rgb}{0.5373, 0.6431, 0.0000}
\definecolor{tu81}{rgb}{0.6784, 0.7490, 0.3020}
\definecolor{tu82}{rgb}{0.7686, 0.8196, 0.4980}
\definecolor{tu83}{rgb}{0.8588, 0.8941, 0.6980}
\definecolor{tu84}{rgb}{0.9059, 0.9294, 0.8000}
\definecolor{tu9}{rgb}{0.0000, 0.4431, 0.3373}
\definecolor{tu91}{rgb}{0.3020, 0.6118, 0.5373}
\definecolor{tu92}{rgb}{0.5490, 0.7490, 0.7020}
\definecolor{tu93}{rgb}{0.7490, 0.8588, 0.8353}
\definecolor{tu94}{rgb}{0.8549, 0.9176, 0.9059}
\definecolor{tu10}{rgb}{0.8000, 0.0000, 0.6000}
\definecolor{tu101}{rgb}{0.8706, 0.3490, 0.7412}
\definecolor{tu102}{rgb}{0.9216, 0.6000, 0.8392}
\definecolor{tu103}{rgb}{0.9608, 0.8000, 0.9216}
\definecolor{tu104}{rgb}{0.9804, 0.8980, 0.9608}
\definecolor{tu110}{rgb}{0.4627, 0.0000, 0.4627}
\definecolor{tu111}{rgb}{0.5961, 0.2510, 0.5961}
\definecolor{tu112}{rgb}{0.7294, 0.4980, 0.7294}
\definecolor{tu113}{rgb}{0.8392, 0.6980, 0.8392}
\definecolor{tu114}{rgb}{0.9216, 0.8510, 0.9216}
\definecolor{tu120}{rgb}{0.4627, 0.0000, 0.3294}
\definecolor{tu121}{rgb}{0.6118, 0.3020, 0.5333}
\definecolor{tu122}{rgb}{0.7569, 0.5490, 0.6980}
\definecolor{tu123}{rgb}{0.8667, 0.7490, 0.8314}
\definecolor{tu124}{rgb}{0.9216, 0.8510, 0.9020}
\definecolor{tu130}{rgb}{0.0314, 0.0314, 0.0314}
\definecolor{tu131}{rgb}{0.3725, 0.3725, 0.3725}
\definecolor{tu132}{rgb}{0.5882, 0.5882, 0.5882}
\definecolor{tu133}{rgb}{0.7529, 0.7529, 0.7529}
\definecolor{tu134}{rgb}{0.8667, 0.8667, 0.8667}
\definecolor{tu140}{rgb}{0.0000, 0.6875, 0.3125}
\definecolor{light-gray}{gray}{0.90}

\usetikzlibrary{arrows.meta, positioning, calc}

\begin{document}

\title{Spatio-Temporal Attention for Consistent Video Semantic Segmentation in Automated Driving}


\author{\authorname{Serin Varghese\sup{1,2}, Kevin Ro\ss\sup{1}, Fabian H\"uger\sup{2} and Kira~Maag\sup{1}}
\affiliation{\sup{1}Heinrich-Heine-University D{\"u}sseldorf, Department of Computer Science, D{\"u}sseldorf, Germany}
\affiliation{\sup{2}CARIAD SE, Wolfsburg, Germany}
\email{\{serin.varghese, fabian.hueger\}@cariad.technology, \{kevin.ross, kira.maag\}@hhu.de}
}


\keywords{Video Semantic Segmentation, Temporal Consistency, Spatio-Temporal Attention, Transformer Networks, Automated Driving, Computer Vision.}

\abstract{Deep neural networks, especially transformer-based architectures, have achieved remarkable success in semantic segmentation for environmental perception. However, existing models process video frames independently, thus failing to leverage temporal consistency, which could significantly improve both accuracy and stability in dynamic scenes. In this work, we propose a Spatio-Temporal Attention (STA) mechanism that extends transformer attention blocks to incorporate multi-frame context, enabling robust temporal feature representations for video semantic segmentation. Our approach modifies standard self-attention to process spatio-temporal feature sequences while maintaining computational efficiency and requiring minimal changes to existing architectures. STA demonstrates broad applicability across diverse transformer architectures and remains effective across both lightweight and larger-scale models. A comprehensive evaluation on the Cityscapes and BDD100k datasets shows substantial improvements of 9.20 percentage points in temporal consistency metrics and up to 1.76 percentage points in mean intersection over union compared to single-frame baselines. These results demonstrate STA as an effective architectural enhancement for video-based semantic segmentation applications.}

\onecolumn \maketitle \normalsize \setcounter{footnote}{0} \vfill

\section{\uppercase{Introduction}}
\label{Chapter:Introduction}

\textbf{Semantic segmentation} has emerged as a cornerstone in the field of computer vision, enabling pixel-level classification of image content into a predefined set of semantic classes for comprehensive scene understanding.

\begin{figure}[t!]
\centering
\begingroup
\renewcommand{\arraystretch}{1.2}
\makebox[\columnwidth]{%
\begin{tabular}{>{\centering\arraybackslash}m{0.3\columnwidth}
                >{\centering\arraybackslash}m{0.3\columnwidth}
                >{\centering\arraybackslash}m{0.3\columnwidth}}
\multicolumn{3}{c}{\textbf{Time instances}} \\
\textit{$t{-}2$} & \textit{$t{-}1$} & \textit{$t$} \\
\includegraphics[width=\linewidth]{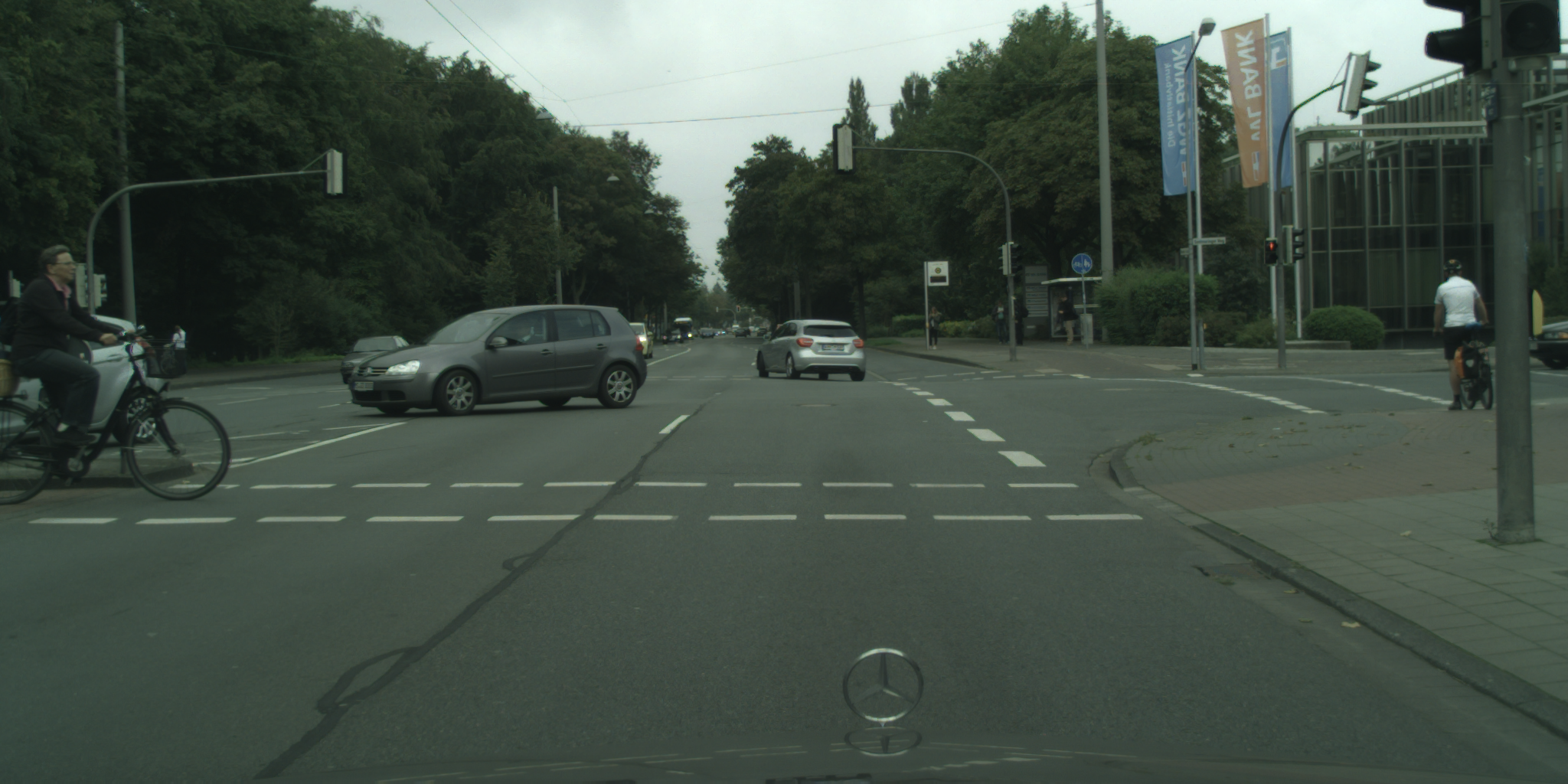} &
\includegraphics[width=\linewidth]{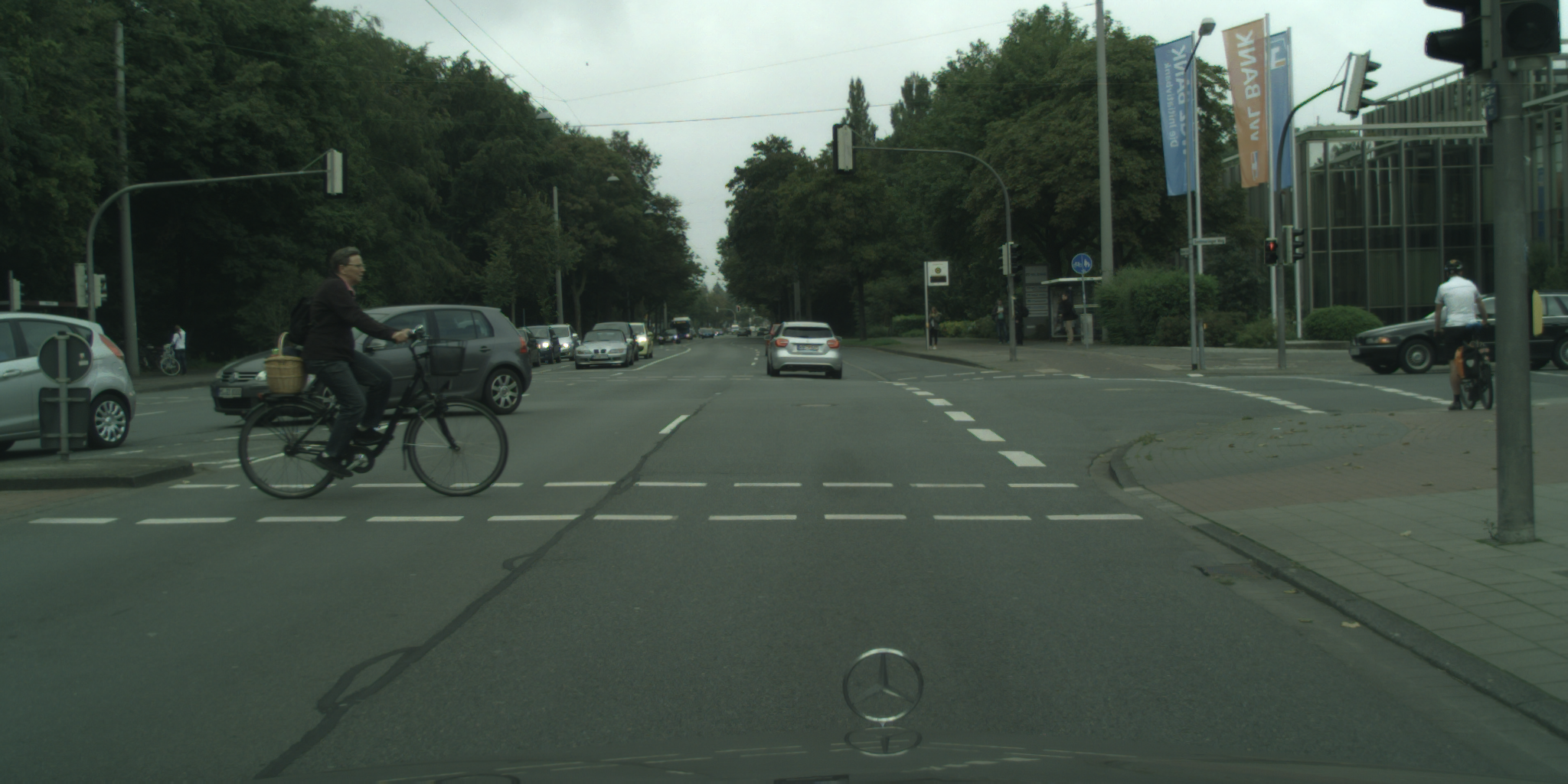} &
\includegraphics[width=\linewidth]{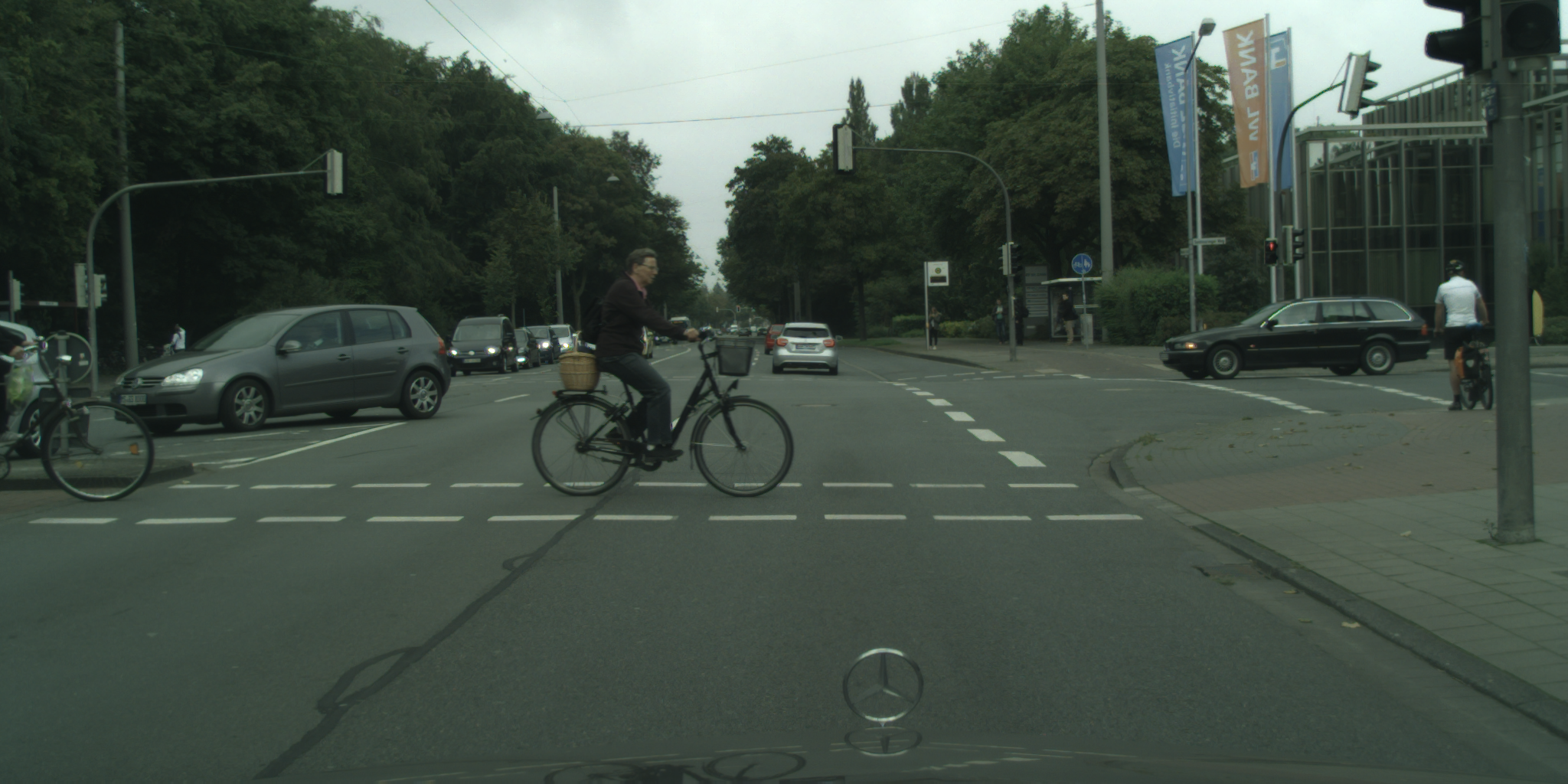} \\
\multicolumn{3}{c}{\textbf{Unstable predictions}} \\
\includegraphics[width=\linewidth]{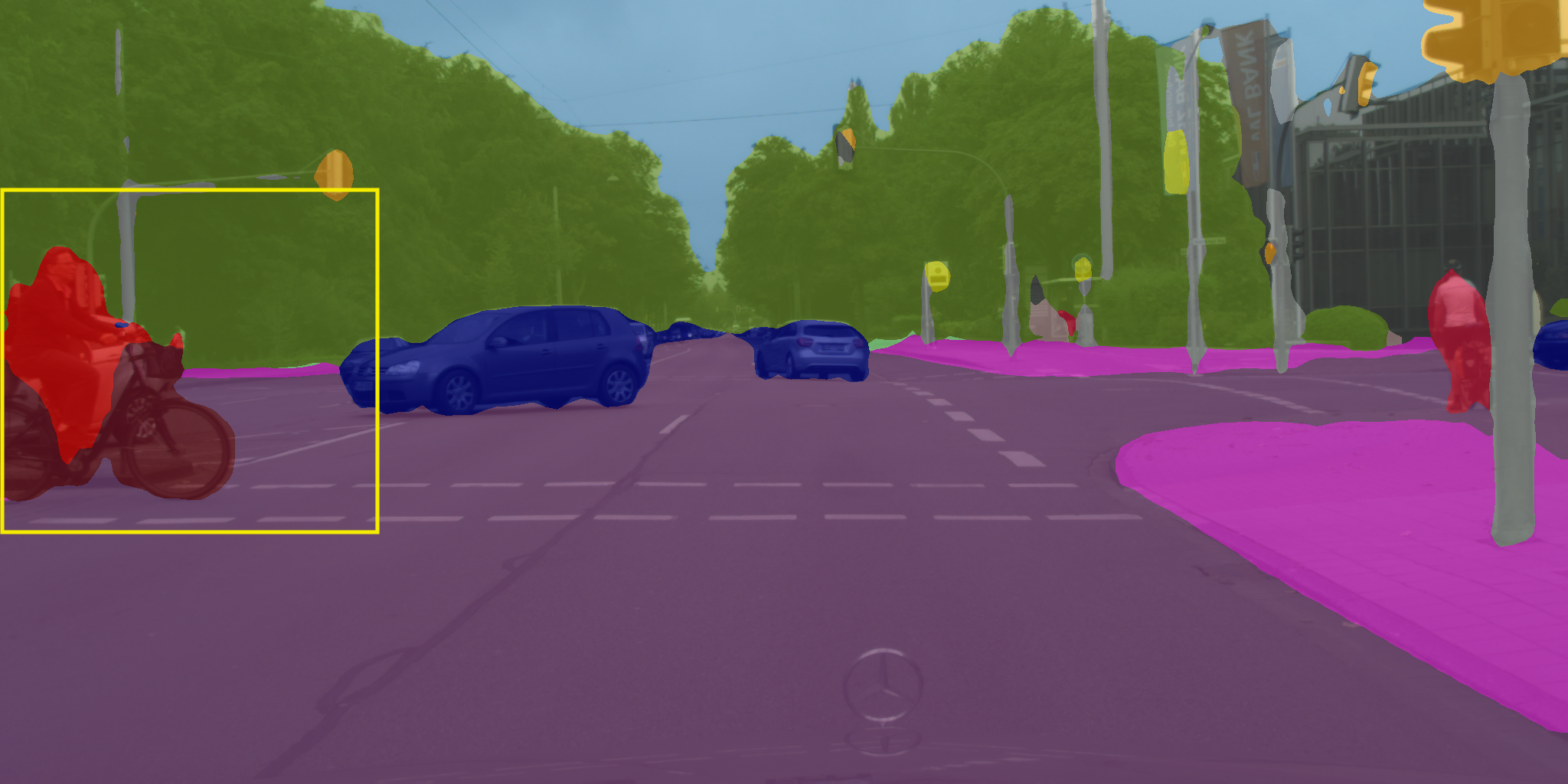} &
\includegraphics[width=\linewidth]{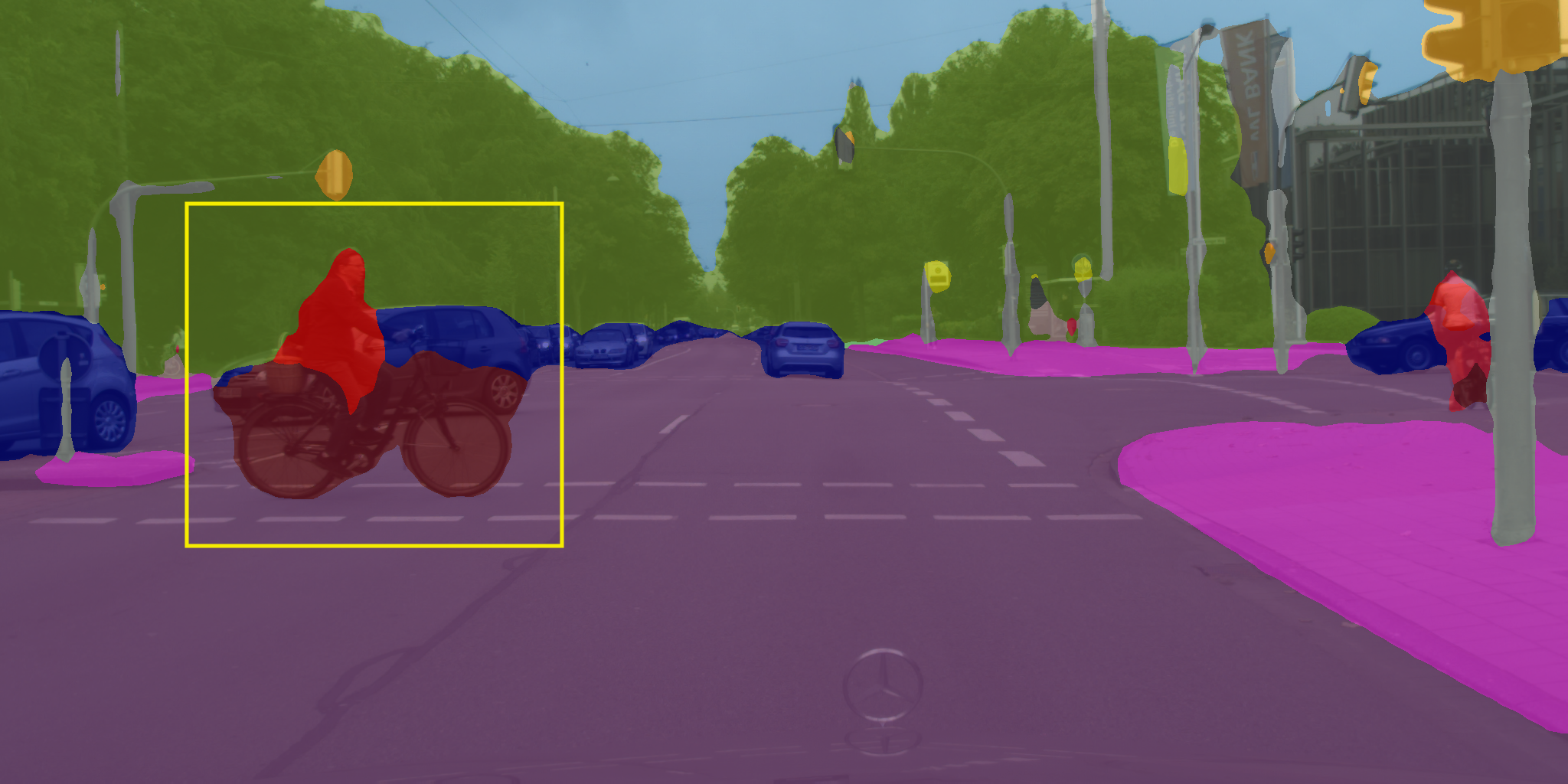} &
\includegraphics[width=\linewidth]{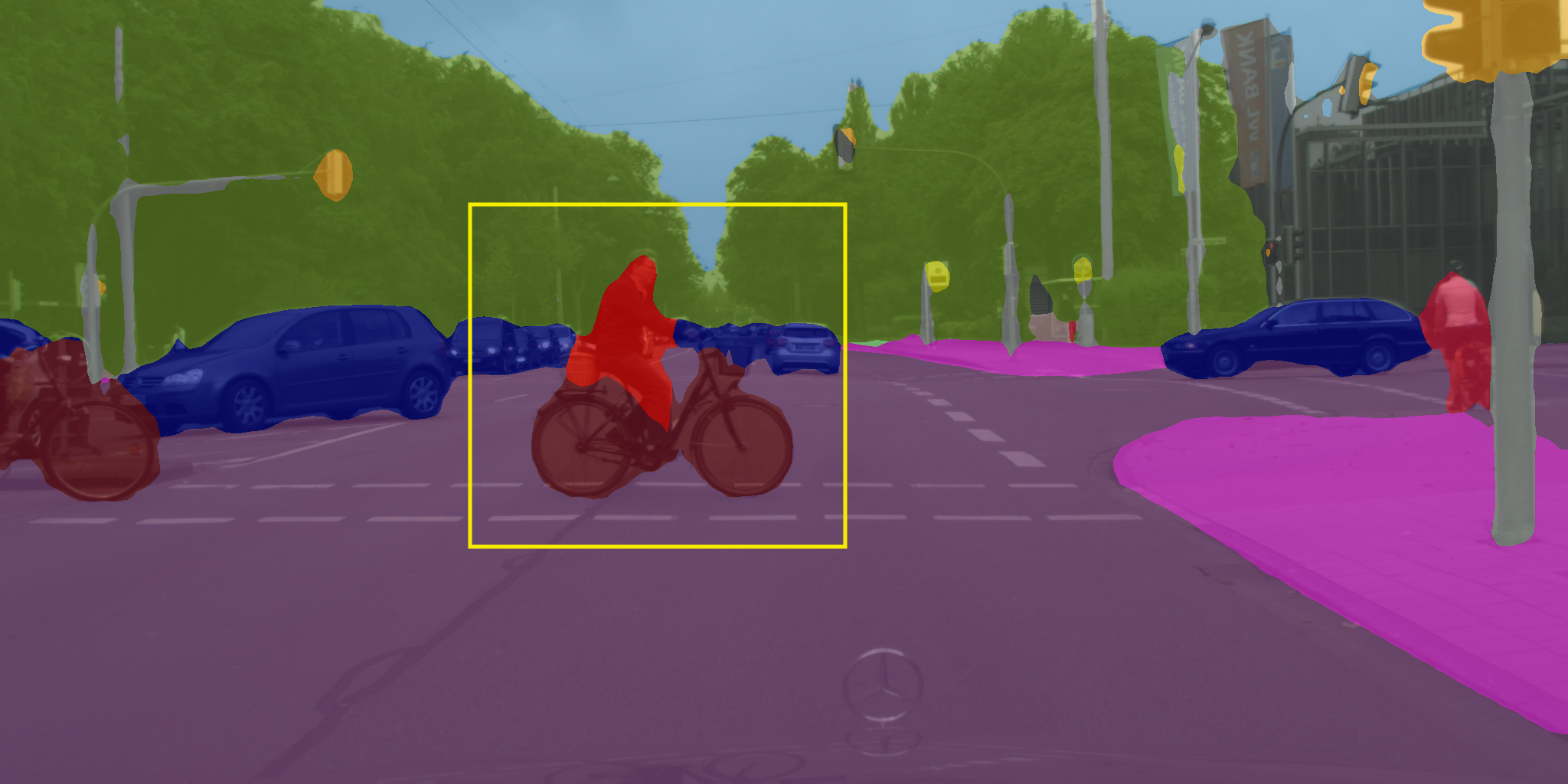} \\
\multicolumn{3}{c}{\textbf{Our method}} \\
\includegraphics[width=\linewidth]{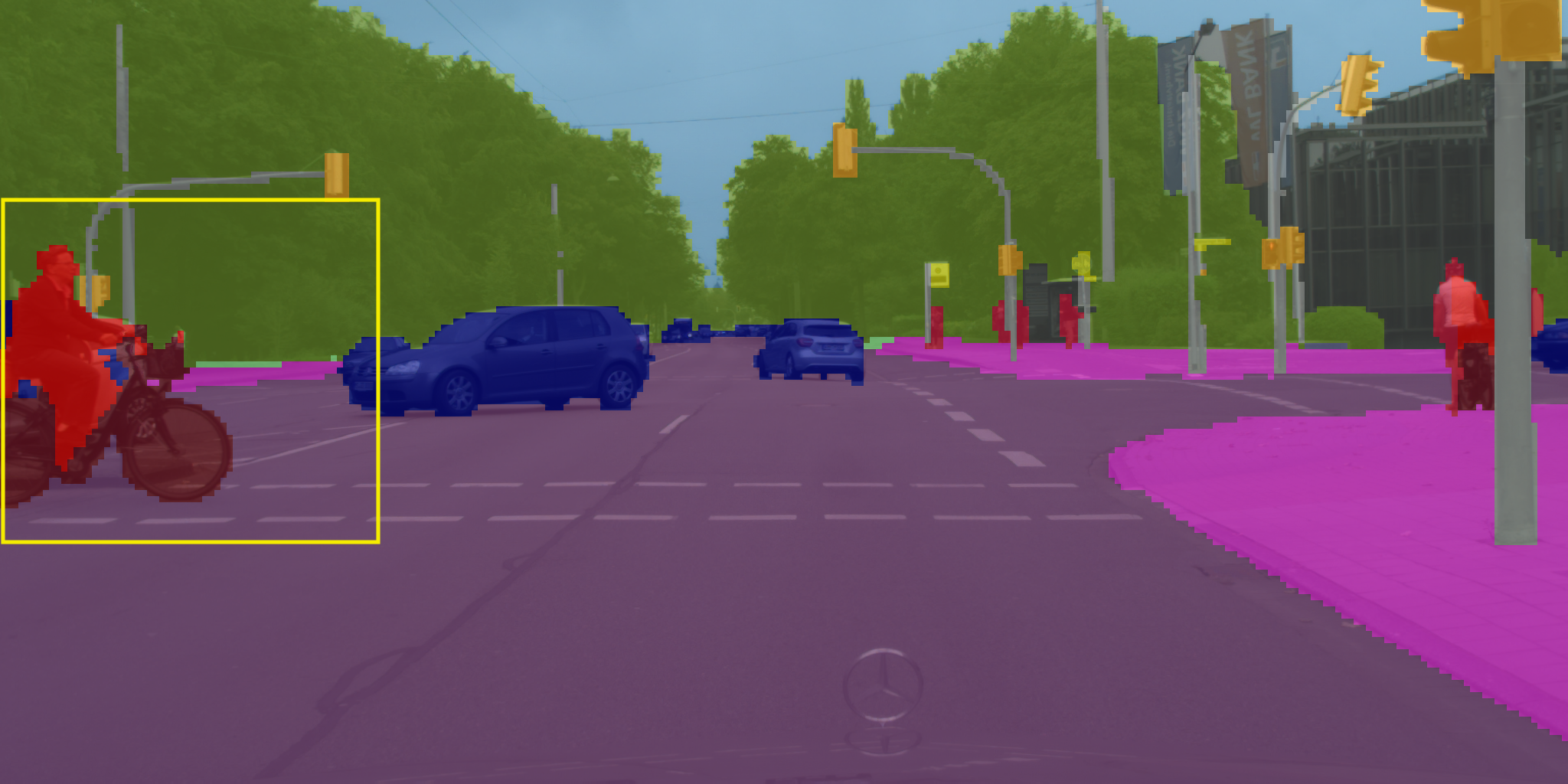} &
\includegraphics[width=\linewidth]{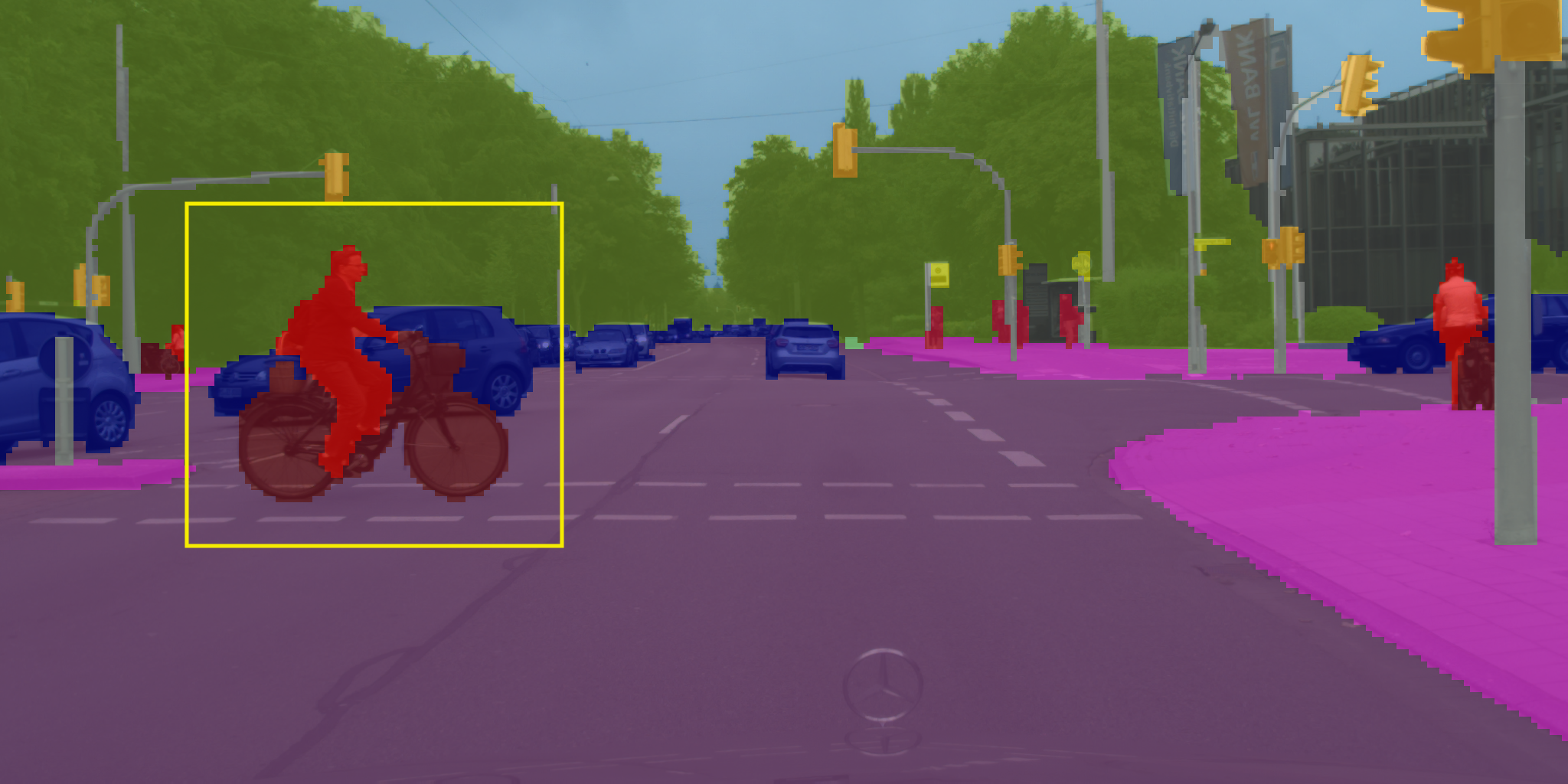} &
\includegraphics[width=\linewidth]{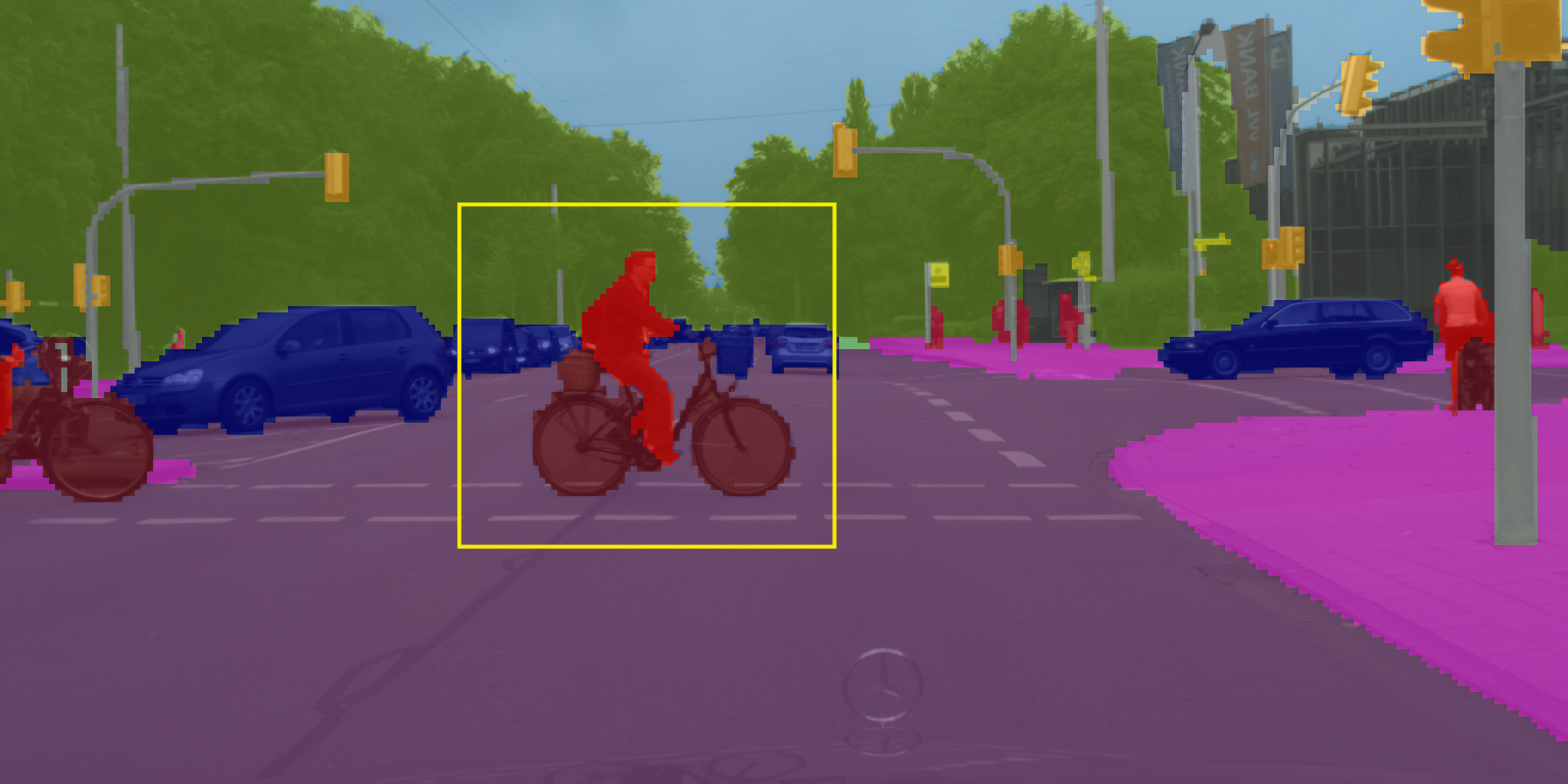} \\
\end{tabular}%
}\endgroup
\caption{
\textbf{Examples of stable and unstable predictions.} The yellow box highlights the area of interest in the images. 
\emph{Top:} From left to right, we have three consecutive frames of a video sequence from $t-2$ to $t$. 
\emph{Center:} The predictions of a semantic segmentation model without our spatio-temporal attention module (STA), where the motorcycle and bicycles are not consistent over time (both $t\!\!-\!\!2\rightarrow t\!\!-\!\!1$, and $t\!\!-\!\!1\rightarrow t$).
\emph{Bottom:} STA focuses on improving temporal consistency of predictions of semantic segmentation networks over time. With our approach we observe an improvement in the robustness of the prediction in the highlighted area.
}
\label{fig:example_image}
\end{figure}
By assigning semantic labels to every pixel, semantic segmentation finds applications in diverse areas such as medical imaging, automated driving, and augmented reality. 
Convolutional neural networks~(CNNs) have long been the backbone of semantic segmentation, with encoder-decoder architectures such as U-Net~\cite{Ronneberger2015}, DeepLabv3+~\cite{Chen2018a} and HRNet~\cite{Wang2019b} achieving strong performance through techniques like skip connections, atrous spatial pyramid pooling and multi-scale feature aggregation. 
More recently, transformer-based models, such as SegFormer~\cite{Xie2021} and SETR~\cite{Zheng2021}, have gained prominence, as their self-attention mechanism enables global context modeling and improved scalability, offering significant advantages over traditional CNNs when processing complex scene structures.

Independent of the model architecture, these methods primarily work with static images and treat each image in isolation. While effective for tasks involving single images, their inability to utilize temporal information contained in sequential data limits their applicability in dynamic environments such as automated driving. In these applications, video data is often available and a reliable, temporally consistent prediction is of utmost interest. 
An example of a single-frame prediction that is unstable over time is shown in Fig.~\ref{fig:example_image}.

\textbf{Temporal consistency} refers to the smoothness and stability of predictions over sequential frames in video data. 
This is an important consideration for applications in safety-critical domains such as robotics and automated driving, where the consistency of predictions is a key factor in ensuring robust system behavior~\cite{Maag2021,Rottmann2020a}.
Challenges in maintaining temporal consistency stem from factors such as occlusions, lighting changes, and rapid scene dynamics. 
In~\cite{Varghese2021}, temporal consistency based on optical flow was defined as the property that objects or structures in an image sequence remain consistently represented over time when their motion is accounted. Optical flow provides the estimated displacement of pixels between two consecutive frames~\cite{Sundaram2010}. If the semantic segmentation prediction of a model from one frame is transferred (i.e., ``warped'') to the next frame using optical flow, the prediction is considered temporally consistent if it matches the actual model prediction in the next frame. Deviations indicate temporal inconsistency, e.g., when boundaries flicker, objects jump, or textures are not stable.

\textbf{Video semantic segmentation} combines the demands of accurate spatial segmentation with the need for temporally coherent predictions across consecutive frames. 
In contrast to image segmentation, video segmentation must additionally address challenges unique to video data, such as motion blur, varying frame rates, and cross-frame feature correspondence. 
As a result, video segmentation requires models that not only capture fine-grained spatial detail but also enforce stability over time. 
Early approaches often relied on motion cues such as optical flow to enforce temporal consistency~\cite{Varghese2021}, but these methods are computationally expensive since they require estimating pixel-wise motion fields across frames. In comparison, recent transformer-based methods typically use 3D attentions~\cite{Bertasius2021}, which have high computational costs, or introduce plug-in modules that extend existing architectures to model temporal coherence~\cite{Li2021b}, offering a more scalable alternative to dense optical flow computation. 
In comparison, we present a first standalone transformer approach for video semantic segmentation that integrates temporal context and can be integrated into common transformer architectures with minimal overhead.

In this work, we propose a novel Spatio-Temporal Attention (STA) mechanism that 
enhances transformer-based video semantic segmentation by directly integrating temporal reasoning into the core attention module. 
In contrast to existing methods, which either depend on computationally expensive optical flow to align features across frames or create new architectures (CNN or transformer) with spatio-temporal components, our approach eliminates the need for explicit motion estimation and can be easily integrated into any transformer-based architectures. 
STA extends the standard formulation of self-attention by incorporating information from (multiple) previous frames into the attention calculation of the current frame. This allows the model to capture cross-frame feature correlations and ensure that the resulting feature maps are enriched with temporal context while preserving fine-grained spatial details. 
As a result, STA not only improves segmentation accuracy but also enforces temporal consistency across consecutive frames, a crucial property in video-based applications such as automated driving. 
The benefit of our method is that we calculate the attention information from the previous frames anyway and can feed them directly into the attention mechanism of the current frame without an additional module, which keeps the additional computational overhead very low. \par

To validate its effectiveness, we integrate STA into state-of-the-art transformer architectures, SegFormer~\cite{Xie2021} and UMixFormer~\cite{Zhou2023}, and conduct a systematic evaluation across different model scales, demonstrating both its general applicability and scalability to varying computational budgets.
Our comprehensive evaluation on Cityscapes~\cite{Cordts2016} and BDD100k~\cite{Yu2018b}
datasets demonstrates substantial improvements in both spatial accuracy and temporal
consistency. 
STA-enhanced models achieve improvements of up to $1.76$ percentage points in mean
intersection over union (mIoU) and remarkable gains of up to $9.20$ percentage points in mean
temporal consistency (mTC), with the most substantial improvements observed on
challenging driving scenarios in BDD100k. The temporal context ablation study
reveals that using two previous frames for the prediction 
provide optimal performance, offering practical guidance
for real-time applications where computational efficiency is paramount.


\section{\uppercase{Related Work}}
\label{Chapter:Related Work}

\textbf{Optical-flow based temporal consistency.} \hspace{1ex}
Temporal consistency is a critical aspect in video semantic segmentation, as inconsistent predictions across frames can lead to perceptual artifacts and unreliable scene understanding~\cite{Maag2019,Maag2021}. 
Early works addressed temporal smoothing using post-processing methods that require semantic segmentation prediction and then refine the predictions using optical flow~\cite{Dong2015,Hur2016}, 
but these approaches often failed to capture long-range dependencies and struggled with dynamic objects. 
Subsequent approaches integrate optical flow directly into the training process rather than as post-processing refinement.
In~\cite{Jain2019}, the predictions of two network branches are combined: a reference branch that extracts highly detailed features on a reference frame and warps these features forward using frame-to-frame optical flow estimates, and an update branch that computes features on the current frame and performs a temporal update for each video frame. 
In~\cite{Zhu2017b}, deep feature flow was proposed, where the expensive convolutional sub-network is executed only on sparse key frames and the resulting deep feature maps are propagated to other frames via a flow field. 
The works of Varghese et al.~\cite{Varghese2021} addressed this problem through specialized loss functions and architectural modifications, advancing the understanding of temporal consistency in automotive applications. 
While these optical flow-based methods provide valuable complementary approaches to temporal modeling, they require separate motion estimation pipelines that can be computationally intensive and error-prone, motivating the development of methods that incorporate temporal reasoning directly within network architectures.

\textbf{Video semantic segmentation.} \hspace{1ex}
Video semantic segmentation has evolved from simple frame-by-frame processing to sophisticated architectures that explicitly model temporal dependencies. 
CNN-based video segmentation methods typically introduce temporal modules such as convolutional long short-term memory (ConvLSTM, \cite{Zhou2021a}) or 3D convolutions \cite{schmidt2021} to incorporate temporal cues. 
In \cite{Shelhamer2016}, ``clockwork'' CNNs are introduced driven by fixed or adaptive clock signals that schedule the processing of different layers at different update rates according to their semantic stability.
However, these CNN-based approaches are fundamentally limited by their localized receptive fields, struggling with long-range temporal dependencies that are crucial for maintaining consistency across video sequences.

In addition, attention modules are used in CNNs. Two examples of such architectures are TDNet \cite{HuZhu2019a}, which uses attention propagation modules to efficiently combine sub-features across frames, and TMANet \cite{Wang2021a}, which considers self-attention to aggregate the relations between consecutive video frames.
Transformer-based approaches have emerged as powerful alternatives, naturally modeling long-range dependencies via self-attention. 
However, standard vision transformers operate purely in the spatial domain on single frames, without incorporating temporal information \cite{Xie2021,Zheng2021}. 
Architectures like TimeSformer~\cite{Bertasius2021}, ViViT~\cite{Arnab2021}, and Video Swin Transformers~\cite{Liu2022c} extend transformers to video understanding by explicitly modeling spatio-temporal dependencies. 
TimeSformer factorizes attention into spatial and temporal components, reducing the quadratic cost of full 3D attention. 
ViViT, in contrast, explores multiple variants of spatio-temporal attention, providing a flexible framework for video modeling. 
Video Swin Transformer builds on the hierarchical Swin architecture by applying shifted windows in both space and time, enabling efficient local spatio-temporal modeling at multiple scales. 
While these architectures demonstrate the potential of transformer-based video understanding, they typically require complete architectural redesigns~\cite{Baghbaderani2024,Weng2023,Xing2022} and substantial computational overhead~\cite{Park2022,Yang2024}, limiting their adaptability to existing transformer models.
An alternative approach involves modular plug-in temporal modules that add temporal reasoning to existing networks without modifying the core architecture.
The Sparse Temporal Transformer (STT, \cite{Li2021b}) introduces a temporal module that captures cross-frame context using query and key selection, which encodes temporal dependencies from previous frames.

Despite these advances, a fundamental gap remains in transformer architectures for video understanding: the lack of a unified approach that integrates temporal reasoning directly into the core attention mechanism while maintaining architectural generalizability. 
Existing methods either require complete architectural redesigns (limiting adaptability), introduce separate temporal modules (adding computational overhead), or rely on external motion estimation (introducing error propagation). 
This gap motivates the need for an architectural enhancement that naturally extends the transformer's attention mechanism to the temporal domain while preserving the flexibility to be integrated across different transformer-based architectures without substantial modifications.
Our STA mechanism addresses this fundamental gap by extending the core self-attention computation to incorporate temporal context, representing a natural evolution of transformer architectures toward unified spatio-temporal reasoning.

\section{\uppercase{Spatio-Temporal Attention}}
\label{Chapter:Method}
\begin{figure*}[ht!]
    \centering
    \resizebox{0.9\linewidth}{!}{
    \input{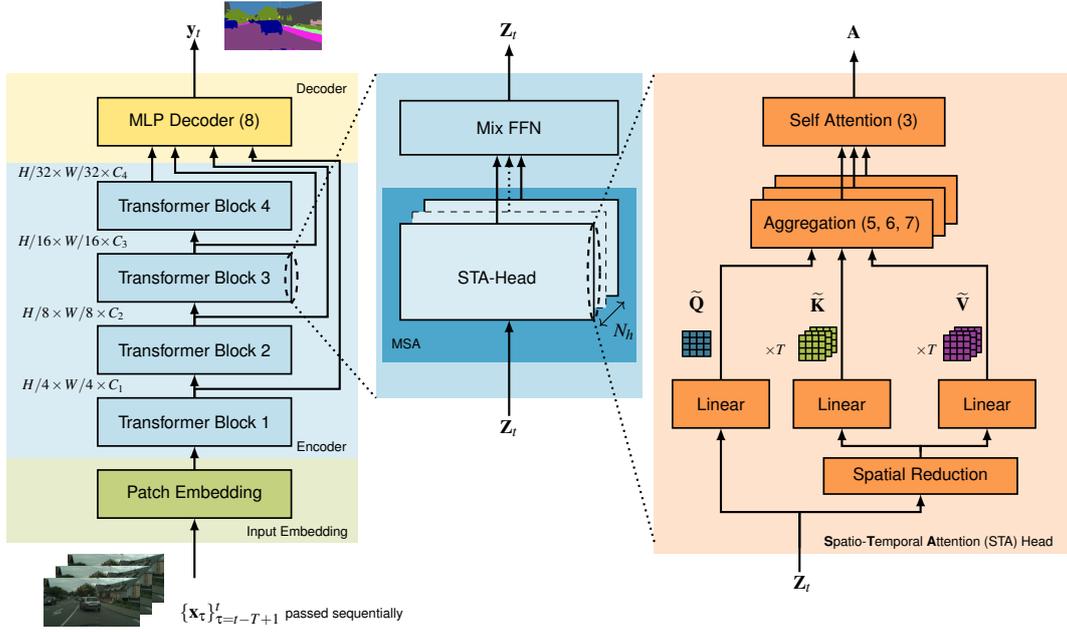} }
    \caption{\textbf{Hierarchical illustration of our proposed Spatio-Temporal Attention (STA) module.} The figure shows three levels of detail.
\emph{Left:} Transformer-based segmentation architecture with MiT encoder stages processing multi-scale features. 
\emph{Center:} Individual transformer block with Multi-Head Self-Attention (MSA) containing multiple STA-Heads, followed by Mix-FFN.  
\emph{Right:} Detailed STA computation that extends standard attention across temporal sequence $\{\mathbf{x}_{t-T+1}, \ldots, \mathbf{x}_t\}$, enabling cross-frame feature aggregation while preserving spatial relationships.}
\label{fig:STA_Method}
\end{figure*}{}

In this section, we detail our proposed approach for enhancing temporal consistency in video semantic segmentation through the introduction of the STA module.

Let $\widetilde{\mathcal{D}} = \{ d \mid d = \{\mathbf{x}_t\}_{t=1}^{N_d} \}$ denote the dataset of video sequences, where each sequence $d$ contains a respective number of $N_d$ frames $\mathbf{x}_t$.  
Here, $\mathbf{x}_t \in \mathbb{R}^{H \times W \times 3}$ denotes the RGB image at time $t$, 
with $H$ and $W$ representing the image height and width, respectively.

Only a subset of these frames is annotated with ground truth labels, so we define the annotated dataset $\mathcal{D} = \{ (\mathbf{x}_t, \mathbf{m}_t) \mid \mathbf{x}_t \in d, \, d \in \widetilde{\mathcal{D}}, \, t \in \mathcal{T}_d \}$ where $\mathbf{m}_t$ denotes the segmentation mask of frame $\mathbf{x}_t$, and $\mathcal{T}_d \subseteq \{1, \dots, N_d\}$ is the set of annotated frame indices within sequence $d$. 
Each segmentation mask $\mathbf{m}_{t} \in \mathcal{L}^{H \times W}$ has a pixel-wise class label $\mathcal{L}~=~\{1,\ldots,C\}$ representing the $C$ semantic classes. 
Since not all frames are labeled in video sequence datasets, pseudo labels are generated for the frames $t \notin \mathcal{T}_d, d \in \widetilde{\mathcal{D}}$, 
ensuring consistent supervision across the temporal sequence. 

Semantic segmentation can be viewed as the task of assigning each pixel of an input frame $\mathbf{x}_t$ a semantic category. 
Given learned weights, a neural network provides a pixel-wise probability distribution $\mathbf{y}_t \in [0,1]^{H \times W \times C}$ as final prediction. 
Transformer-based architectures model the global dependencies through self-attention \cite{Vaswani2017}. Instead of processing the full image, the input frame is first divided into patches and embedded into a sequence of tokens that serve as the input to the transformer layers.

Figure~\ref{fig:STA_Method} provides a comprehensive overview of our approach, illustrating the patch embedding process, transformer architecture, and detailed STA mechanism that we now describe.
The input image $\mathbf{x}_t$ is divided into non-overlapping patches of size $P \times P$, resulting in $L = \frac{H \cdot W}{P^2}$ patches per image. 
Each patch $\mathbf{p}_{t,l} \in \mathbb{R}^{P \times P \times 3}, l=1,\ldots,L, $ is flattened and linearly projected into the $h$-dimensional embedding space
\begin{align}
    \mathbf{z}_{t,l} = \mathrm{Linear}(\mathrm{Flatten}(\mathbf{p}_{t,l})) + \mathbf{e}_{pos,l} \in \mathbb{R}^h \enspace,
\end{align}
where $\mathrm{Linear}(\cdot)$ represents a learnable projection matrix and $\mathbf{e}_{pos,l} \in \mathbb{R}^h$ is the spatial positional encoding for patch position $l$. 
The resulting patch embeddings form the feature representation $\mathbf{Z}_t = [\mathbf{z}_{t,1}, \mathbf{z}_{t,2}, \ldots, \mathbf{z}_{t,L}] \in \mathbb{R}^{L \times h}$ for frame $t$ that is then passed to the transformer encoder blocks.

The first transformer block receives the patch embeddings $\mathbf{Z}_t$ as input and employs multi-head self-attention with $N_h$ STA-Heads. Within each STA-Head, the input features undergo the spatio-temporal attention computation described below, replacing standard spatial-only attention with our temporal-aware mechanism.
Within each attention head, the input features undergo linear transformations to generate the query matrix $\mathbf{Q}_t = \mathbf{Z}_t \mathbf{W}_Q$, key matrix $\mathbf{K}_t = \mathbf{Z}_t \mathbf{W}_K$, and value matrix $\mathbf{V}_t = \mathbf{Z}_t \mathbf{W}_V$, where $\mathbf{W}_Q, \mathbf{W}_K, \mathbf{W}_V \in \mathbb{R}^{h \times h_{\textrm{head}}}$ represent the learnable projection matrices with $h_{\textrm{head}} = h/N_h$ dimension per head.
The self-attention per head is calculated as 
\begin{equation}
    \mathbf{A} = \mathrm{Softmax}\!\left(\frac{\mathbf{Q}\mathbf{K}^\top}{\sqrt{h_{\textrm{head}}}}\right)\mathbf{V} \enspace.
    \label{eq:Attention}
\end{equation}
The results of all heads are then concatenated and combined using a linear projection 
\begin{equation}
    \mathrm{MSA}(\mathbf{Z}_t) = \mathrm{Concat}(\mathbf{A}_1, \dots, \mathbf{A}_{N_h}) \mathbf{W}_O \enspace,
    \label{eq:MSA}
\end{equation}
where $\mathbf{W}_O \in \mathbb{R}^{h \times h}$ is a projection onto the original embedding space. 

An encoder block combines $\mathrm{MSA}$ with a feed-forward network ($\mathrm{FFN}$), using residual connections and layer normalization
\begin{align}
    \mathbf{Z}_t & = \mathrm{LayerNorm}\big(\mathbf{Z}_t + \mathrm{MSA}(\mathbf{Z}_t)\big) \notag \\ 
    \mathbf{Z}_t & = \mathrm{LayerNorm}\big(\mathbf{Z}_t + \mathrm{FFN}(\mathbf{Z}_t)\big) \enspace.
\end{align}

The resulting transformed features $\mathbf{Z}_t$ are passed to both the decoder and the subsequent transformer block, as illustrated in the hierarchical overview.

Our \textbf{Spatio-Temporal Attention (STA)} shown in Figure~\ref{fig:STA_Method} (right panel) extends standard transformer attention to incorporate temporal context across video frames.  
For temporal context length $T$, i.e., the number of previous frames we use to predict frame $t$, we process frames $\{\mathbf{x}_{t-T+1}, \ldots, \mathbf{x}_{t-1}, \mathbf{x}_t\}$ sequentially through the transformer encoder to obtain query, key, and value matrices $\{\mathbf{Q}_\tau, \mathbf{K}_\tau, \mathbf{V}_\tau\}_{\tau=t-T+1}^t$.
Note, if $T=1$, this means that we only consider frame $t$ and do not use any information from previous frames. 
Each frame $\tau$ is processed independently through the transformer encoder up to the deeper layers, generating the corresponding query, key, and value matrices $\mathbf{Q}_\tau$, $\mathbf{K}_\tau$, and $\mathbf{V}_\tau$ that are combined for temporal attention. Since we are dealing with video sequences, we infer the previous frames automatically and store them.

The temporal fusion combines these matrices using exponential weighting to prioritize recent frames:

\begin{align}
\widetilde{\mathbf{Q}} &= \mathbf{Q}_t \label{eq:sta_query} \\
\widetilde{\mathbf{K}} &= \sum_{\tau=t-T+1}^{t} \lambda^{t-\tau} \mathbf{K}_\tau \label{eq:sta_key} \\
\widetilde{\mathbf{V}} &= \sum_{\tau=t-T+1}^{t} \lambda^{t-\tau} \mathbf{V}_\tau \label{eq:sta_value}
\end{align}

where $\lambda \in (0,1]$ is the temporal decay factor, ensuring temporal recency bias.

This asymmetric formulation ensures that the current frame $\mathbf{x}_t$ drives the attention mechanism through its query representation $\mathbf{Q}_t$, while allowing it to selectively attend to relevant spatio-temporal information from previous frames through the aggregated keys and values. 
The current frame thus acts as an \textit{anchor} that determines what temporal information is most relevant for accurate segmentation, rather than allowing past frames to potentially mislead current predictions.

The patch-level features $\mathbf{Z}_t \in \mathbb{R}^{L \times h}$ are spatially upsampled from resolution $L = \frac{H \cdot W}{P^2}$ back to the original image dimensions $H \times W$, following the MLP decoder part (Figure~\ref{fig:STA_Method}, left panel, top). 
Features from different transformer stages are combined using skip connections and hierarchical fusion modules. A lightweight MLP decoder processes the upsampled features to generate per-pixel logits for all $C$ semantic classes
\begin{align}
    \mathbf{L}_t = \mathrm{MLP}(\text{Upsample}(\mathbf{Z}_t)) \in \mathbb{R}^{H \times W \times C} \enspace.
    \label{eq:MLP_decoder}
\end{align}
The softmax normalization produces the final pixel-wise probability distribution
\begin{align}
    \mathbf{y}_t = \textrm{Softmax}(\mathbf{L}_t) \in [0,1]^{H \times W \times C} \enspace,
\end{align}
where $\mathbf{y}_t = (y_{t,1}, y_{t,2}, \dots, y_{t,C})$, 
and the predicted class is then obtained by $\hat{\mathbf{y}}_t = \arg\max_{c \in \mathcal{L}} y_{t,c}$.

\section{\uppercase{Experimental Setting}}
\label{Chapter:Experimental_Setup}

In this section, we describe the experimental setup for evaluating our proposed STA mechanism. First, we present the road-scene datasets used in our experiments, followed by the state-of-the-art transformer-based semantic segmentation networks that serve as our backbone architectures. Finally, we detail the evaluation metrics, hyperparameter settings, and baseline configurations used to assess the performance of our proposed models.

\subsection{Datasets}
\label{subsection:datasets} 
We utilize two widely adopted street-scene datasets, Cityscapes~\cite{Cordts2016} and BDD100k~\cite{Yu2018b}, to evaluate the temporal consistency and segmentation performance of our proposed models. 

\textbf{Cityscapes} is a commonly used dataset for semantic segmentation of dense urban traffic in $50$ different German cities. 
It contains high-resolution images of size $2,\!048 \times 1,\!024$ with pixel-level annotations for various object classes, making it suitable for training and evaluating segmentation models. 
We denote the annotated dataset with high-quality labels by $\mathcal{D}_{\text{CS}}$ consisting of frames which are not temporally related. 
We use the standard splits, i.e., the training set $\mathcal{D}^{\text{train}}_{\text{CS}}$ with $2,\!975$ images, and due to test set upload restrictions, we partition the official validation set into a mini validation set $\mathcal{D}^{\text{val}}_{\text{CS}}$ (Lindau, 59 images) and a test set $\mathcal{D}^{\text{test}}_{\text{CS}}$ (Frankfurt and M\"{u}nster, 441 images), following common practice in prior works~\cite{Nilson2018}. For all baseline models,
only these single annotated frames
are used for training, without leveraging any temporal information. 

Let $\widetilde{\mathcal{D}}_{\text{CS}}$ denote the set of all $5,\!000$ video sequences in Cityscapes, where each sequence $d \in \widetilde{\mathcal{D}}_{\text{CS}}$ contains $30$ frames, $d~=~\{\mathbf{x}_t\}_{t=1}^{30}$ with a frame rate of $17$ frames per second. 
In each of these sequences, there is one annotated frame that is included in the above-mentioned labeled dataset $\mathcal{D}_{\text{CS}}$. 
We split the sequences into training $\widetilde{\mathcal{D}}^{\text{train}}_{\text{CS}}$ and test sequences $\widetilde{\mathcal{D}}^{\text{test}}_{\text{CS}}$, exactly according to the division of the dataset with annotated frames. 
For models utilizing temporal context, we construct temporal sequences by incorporating previous frames from the same video sequence, e.g. for temporal context length $T=3$, we use frames at positions $t-2$, $t-1$, and $t$, where $t$ is the index of the image we aim to predict. 
The corresponding segmentation masks for the temporal context frames ($t-2$ and $t-1$) are created using pseudo labels generated by the pre-trained\footnote{https://github.com/open-mmlab/mmsegmentation, $81.92\%$ mIoU on Cityscapes validation set} strong HRNetV2p-W48~\cite{Wang2019b}, ensuring consistent supervision across the temporal sequence. 
We use the annotated frames and pseudo labels for baseline (single-frame) training as well as temporal training, i.e., same amount of training data for a fair comparison.

\textbf{BDD100k} (Berkeley DeepDrive) contains diverse driving scene videos with a resolution of 1,280$\times$720 pixels. 
Let $\widetilde{\mathcal{D}}_{\text{BDD}}$ denote the set of all 100K video sequences, each with a length of 40 seconds and 30 frames per second. 
10K images are annotated for semantic segmentation, which appear in varying proportions in all sequences. 
We follow the standard splits, i.e., $7,\!000$ training ($\mathcal{D}^{\text{train}}_{\text{BDD}}$), $1,\!000$ validation ($\mathcal{D}^{\text{val}}_{\text{BDD}}$), and $2,\!000$ testing images ($\mathcal{D}^{\text{test}}_{\text{BDD}}$). 
Equivalent to the Cityscapes dataset, we create pseudo labels for the remaining masks in the sequences and maintain the data split there as well. Note, we only generate pseudo labels for the $T-1$ previous frames of $t$ if $t$ has an associated ground truth mask.

\subsection{Semantic Segmentation Models}
\textbf{SegFormer}~\cite{Xie2021} is a semantic segmentation framework that combines transformer-based encoders with lightweight multilayer perceptron~(MLP) decoders. It introduces a hierarchical encoder that produces multiscale feature representations without relying on positional encodings, which improves generalization to varying image resolutions. 
The simple MLP decoder avoids the complexity of traditional segmentation decoders while maintaining high efficiency and accuracy. 
We evaluate two model sizes: \emph{SegFormer-B0} (compact, 3.8M parameters) and \emph{SegFormer-B3} (large, 47M parameters) to assess scalability across different computational budgets.

\textbf{UMixFormer}~\cite{Zhou2023} builds upon the SegFormer framework by integrating a U-shaped encoder-decoder structure with enhanced mixing modules. 
It incorporates skip connections between encoder and decoder stages for richer feature fusion and improved spatial detail recovery. 
The mixing modules enhance the model's ability to capture both local fine-grained details and global contextual information, making it particularly effective for boundary delineation tasks.
We evaluate \emph{UMixFormer-B0} (compact, 5.9M parameters) and \emph{UMixFormer-B3} (large, 58.4M parameters) variants to ensure comprehensive assessment of our STA mechanism across different model scales.

\subsection{Evaluation Metrics}

\textbf{Mean intersection over union} (mIoU) is the primary evaluation metric for semantic segmentation, that calculates the ratio of intersection to union between predicted pixels and ground truth pixels for each class and then averages across all classes.
The mIoU is calculated on individual frames, i.e., each evaluation considers the image independently, without taking into account temporal consistency between consecutive frames.

\textbf{Mean temporal consistency} (mTC)~\cite{Varghese2020} is an unsupervised (label-free) evaluation metric for measuring the stability of the semantic segmentation networks.  
The instantaneous temporal consistency TC$_t$ at time $t$ is defined as 
\begin{equation}
    \text{TC}_t = \text{mIoU}
    (\hat{\mathbf{y}}_t,\Tilde{\mathbf{y}}_t ) \enspace,
    \label{equation: metric mean temporal consistency_2}
\end{equation}
where $\hat{\mathbf{y}}_t$ is the semantic segmentation prediction for frame $t$ and $\Tilde{\mathbf{y}}_t$ is the warped prediction using optical flow computation \cite{Sundaram2010}, i.e., the expected prediction computed based on the prediction of the network at time $t\!\!-\!\!1$ and the movement of the pixels between time $t\!\!-\!\!1$ and $t$.
The mean temporal consistency is then defined as 
\begin{equation}
    \text{mTC} = \frac{1}{N\!-\!1} \sum_{t=2}^{N} \text{TC}_t,
    \label{equation: metric mean temporal consistency}
\end{equation}
where $N$ is the number of frames in the considered sequence.

For evaluation, we use the mIoU to assess the annotated datasets $\mathcal{D}^{\text{test}}_{\#}, \# \in \{ \text{CS}, \text{BDD} \}$ and the mTC for the video sequences $\widetilde{\mathcal{D}}^{\text{test}}_{\#}, \# \in \{ \text{CS}, \text{BDD} \}$.

\subsection{Hyperparameter}

We set the temporal decay factor 
within our STA module to $\lambda\!=\!0.8$, which controls the relative importance of temporal context frames in the attention computation. This value was selected based on empirical validation across a range of settings from $\lambda\!\!\!=\!\!\!0.5$ to $\lambda\!\!=\!\!0.95$, where we observed robust performance within this range. 
The choice 
ensures that recent frames contribute meaningfully to the current prediction while preventing temporal information from overwhelming spatial features.
All experiments are conducted with a temporal context of $T\!\!=\!\!3$ frames, meaning each model processes the current frame along with two preceding frames to generate spatially and temporally consistent segmentation masks. 
In an ablation study, we show that this value is a reasonable compromise between performance and prediction overhead. 
We denote the networks, which integrate our STA mechanism into the backbone architecture, with STA-SegFormer and STA-UMixFormer, respectively.

\subsection{Baselines}
To evaluate the effectiveness of our proposed STA module, we focus on transformer-based architectures (SegFormer and UMixFormer) that have demonstrated strong accuracy and efficiency while being scalable to lightweight variants suitable for resource-constrained applications. We integrate our STA mechanism into both architectures, creating STA-SegFormer and STA-UMixFormer variants that incorporate temporal context processing while preserving original spatial segmentation capabilities. Our primary baselines are the original single-frame models (SegFormer B0/B3 and UMixFormer B0/B3) which process each frame independently without temporal information.

Alternative video segmentation approaches using specialized temporal modules (e.g., ConvLSTMs, 3D convolutions) require significantly higher computational effort and represent standalone architectures rather than extensions of existing models, making them outside the scope of our efficiency-focused evaluation.

\section{\uppercase{Results}}
\label{Chapter:Results}

\newcommand{\ours}{\textcolor{blue}{\textbf{★}}}
\begin{table}[t]
\centering
\caption{\textbf{Quantitative comparison of baselines and STA-enhanced architectures} using temporal context $T=3$. Results show mIoU/mTC (\%) for spatial accuracy and temporal consistency across model sizes. 
Best results are bold.}
\label{table:main_results}
\scalebox{0.94}{
\begin{tabular}{llcc}
\hline
 & & Cityscapes & BDD100k \\
Method & Size & mIoU/mTC & mIoU/mTC \\
\hline
SegFormer & B0 & $72.81$/$75.72$ & $51.40$/$71.40$ \\
\rowcolor{gray!10}
STA-SegFormer & B0 & $\mathbf{74.31\text{/}77.08}$ & $\mathbf{51.64\text{/}80.60}$ \\
\hline
SegFormer & B3 & $80.55$/$76.05$ & $59.79$/$84.74$ \\
\rowcolor{gray!10}
STA-SegFormer & B3 & $\mathbf{81.93\text{/}77.37}$ & $\mathbf{61.55\text{/}88.28}$ \\
\hline
UMixFormer & B0 & $75.12$/$80.91$ & $54.31$/$74.62$ \\
\rowcolor{gray!10}
STA-UMixFormer & B0 & $\mathbf{76.00\text{/}81.38}$ & $\mathbf{55.38\text{/}77.41}$ \\
\hline
UMixFormer & B3 & $82.45$/$82.06$ & $61.88$/$86.45$ \\
\rowcolor{gray!10}
STA-UMixFormer & B3 & $\mathbf{83.82\text{/}84.67}$ & $\mathbf{62.90\text{/}89.63}$ \\
\hline
\end{tabular}

}
\end{table}

We present comprehensive experimental results demonstrating the effectiveness of our proposed spatio-temporal attention mechanism in enhancing temporal consistency for video semantic segmentation. 

\subsection{Quantitative Results}

Table~\ref{table:main_results} presents the main quantitative comparison between baseline architectures and their STA-enhanced counterparts. Note, all models use the same amount of training images.
The results clearly demonstrate the effectiveness of incorporating temporal information through our STA mechanism, with consistent improvements in both spatial accuracy (mIoU) and temporal consistency (mTC) across all evaluated scenarios.

The STA-enhanced models consistently outperform their baseline counterparts across both datasets and model sizes. On Cityscapes, STA-SegFormer B0 achieves a $1.50$ percentage points (pp) improvement in mIoU and a $1.36$ pp gain in temporal consistency compared to the baseline SegFormer B0.
In particular, we are also enhancing the performance of the larger models with STA-SegFormer B3 demonstrating a $1.38$ pp increase in mIoU and $1.32$ pp in mTC. 
Similar trends are observed for the comparatively stronger UMixFormer, where STA-UMixFormer B0 shows a $0.88$ pp improvement in mIoU and $0.47$ pp in mTC, while the B3 variant achieves $1.37$ pp and $2.61$ pp enhancements respectively.

Particularly noteworthy are the substantial improvements in temporal consistency on the BDD100k dataset. STA-SegFormer B0 achieves a remarkable $9.20$ pp performance increase in mTC while maintaining spatial accuracy, and STA-SegFormer B3 demonstrates a $1.76$ pp improvement in mIoU and $3.54$ pp in mTC. These results highlight the significant impact of explicit temporal modeling on both spatial accuracy and temporal stability, with the benefits being more pronounced on challenging driving scenarios represented in BDD100k.

\subsection{Temporal Context Ablation Study}

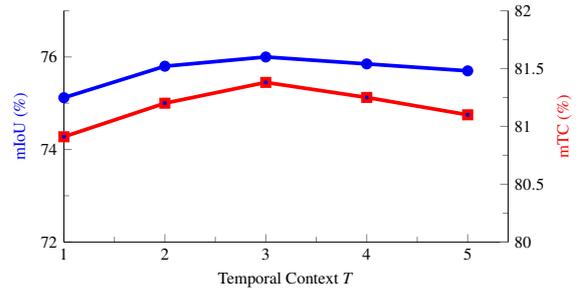
\begin{figure}[t!]
  \centering
  \resizebox{\columnwidth}{!}{%
  \begin{tikzpicture}
    \begin{axis}[
      width=10cm,
      height=6cm,
      xlabel={Temporal Context $T$},
      ylabel={mIoU (\%)},
      ylabel style={blue},
      xtick={1,2,3,4,5},
      xmin=1,
      ymin=72, ymax=77,
      tick label style={font=\small},
      label style={font=\small},
      axis lines*=left,
      minor x tick num=1,
      minor y tick num=1
    ]
      \addplot+[mark=*,line width=2pt,blue] coordinates {
        (1, 75.12)
        (2, 75.80)
        (3, 76.00)
        (4, 75.85)
        (5, 75.70)
      };
    \end{axis}
    
    \begin{axis}[
      width=10cm,
      height=6cm,
      xlabel={Temporal Context $T$},
      ylabel={mTC (\%)},
      ylabel style={red},
      xtick={1,2,3,4,5},
      xmin=1,
      ymin=80, ymax=82,
      axis y line*=right,
      axis x line=none,
      tick label style={font=\small},
      label style={font=\small},
      minor x tick num=1,
      minor y tick num=1,
      legend style={at={(0.3,0.85)},anchor=south,legend columns=-1}
    ]
      \addplot+[mark=square*,line width=2pt,red] coordinates {
        (1, 80.91)
        (2, 81.20)
        (3, 81.38)
        (4, 81.25)
        (5, 81.10)
      };
    \end{axis}
  \end{tikzpicture}%
  }
  \caption{\textbf{Temporal context ablation study} showing performance of STA-UMixFormer B0 on the Cityscapes dataset across different temporal context lengths $T$. The value of $T=1$ corresponds to the single frame procedure.}
  \label{fig:temporal_ablation}
\end{figure}

To understand the effect of temporal context length on model performance, we conduct an ablation study varying the number of previous frames considered during training and inference. Fig.~\ref{fig:temporal_ablation} shows the results (mIoU and mTC) for STA-UMixFormer B0 on Cityscapes, where we increase the temporal context from $T=1$ (single frame) to $T=5$ (current frame plus four previous frames).
We made similar observations for the other dataset and architectures.


The ablation study reveals several key insights. First, incorporating even minimal temporal context ($T\!\!\!~=~\!\!\!2$) provides significant improvements over single-frame processing, with mTC showing the largest increase. 
The optimal performance is achieved at $T\!\!=\!\!3$, where both spatial accuracy and temporal consistency reach their peak values. 
This aligns with our asymmetric attention design where the current frame queries can effectively leverage information from 2-3 previous frames without being overwhelmed by excessive temporal context.
Beyond $T\!\!=\!\!3$, we observe diminishing returns and occasionally slight performance degradation, confirming that our current-frame-as-anchor approach prevents the model from over-relying on potentially outdated temporal information.

The choice of temporal context length \(T\) is intrinsically linked to scene dynamics in automated driving. At Cityscapes’ 17~fps frame rate, \(T=3\) covers approximately \(0.18\) seconds of temporal history, which is sufficient to capture typical object motions while avoiding outdated information. Larger values of \(T\) risk incorporating spatially displaced features that no longer correspond to current object positions, particularly for fast-moving objects. Our exponential decay factor \(\lambda\) mitigates motion discontinuities—such as sudden occlusions, appearance changes, or direction shifts—by down-weighting older frames while still leveraging their contextual information. The ablation study (Fig.~\ref{fig:temporal_ablation}) empirically confirms that \(T>3\) degrades performance, as the model begins incorporating features from positions where objects have significantly displaced or disappeared, introducing noise rather than useful temporal context. This temporal window represents an effective balance between capturing smooth motion patterns and avoiding stale information in dynamic driving scenarios.

As expected, the temporal consistency metric shows greater sensitivity to context length compared to spatial accuracy. This suggests that while the quality of spatial segmentation benefits slightly from additional temporal information, the stability of predictions across multiple frames depends largely on the appropriate temporal context window.

\subsection{Computational Overhead Analysis}

To evaluate the practical feasibility of STA for real-time automated driving applications, we conduct a comprehensive computational overhead analysis on NVIDIA RTX 6000 GPUs. Table~\ref{table:computational_overhead} presents detailed measurements of FLOPs, size, and frame per second~(FPS) for both baseline and STA-enhanced models.

\begin{table}[t!]
\centering
\caption{\textbf{Computational overhead analysis} for STA-enhanced models with temporal context $T=3$ on NVIDIA RTX 6000. Values measured on Cityscapes with batch size 1. FLOPs are specified in giga, model size in million parameters. 
}
\label{table:computational_overhead}
\scalebox{0.94}{
\begin{tabular}{lccc}
\toprule
Method & FLOPs $\downarrow$ & Size $\downarrow$ & FPS $\uparrow$ \\
\midrule
SegFormer B0 & $15.7$ & $3.8$ & $28.5$ \\
\rowcolor{gray!10}
STA-SegFormer B0 & $18.5$ & $4.2$ & $24.2$ \\
\textit{Overhead} & \textit{+18\%} & \textit{+11\%} & \textit{-15\%} \\
\midrule
SegFormer B3 & $79.1$ & $47.0$ & $9.2$ \\
\rowcolor{gray!10}
STA-SegFormer B3 & $97.3$ & $53.8$ & $7.6$ \\
\textit{Overhead} & \textit{+23\%} & \textit{+14\%} & \textit{-17\%} \\
\midrule
UMixFormer B0 & $18.9$ & $5.9$ & $24.1$ \\
\rowcolor{gray!10}
STA-UMixFormer B0 & $22.3$ & $6.6$ & $20.5$ \\
\textit{Overhead} & \textit{+18\%} & \textit{+12\%} & \textit{-15\%} \\
\midrule
UMixFormer B3 & $95.2$ & $58.4$ & $7.8$ \\
\rowcolor{gray!10}
STA-UMixFormer B3 & $117.0$ & $66.7$ & $6.5$ \\
\textit{Overhead} & \textit{+23\%} & \textit{+14\%} & \textit{-17\%} \\
\bottomrule
\end{tabular} }
\end{table}

The STA mechanism introduces a moderate overhead of approximately $18-23\%$ in FLOPs across all model configurations, with B0 variants showing lower overhead ($18\%$) compared to B3 models ($23\%$). This overhead is reasonable, considering the substantial performance improvements achieved, particularly in temporal consistency. 
Parameter overhead remains modest at $11$-$14\%$, confirming that STA enhances model capability primarily through architectural improvements rather than brute-force parameter scaling. 

The inference times of $24.2$ FPS for SegFormer B0 and $20.5$ FPS for UMixFormer B0 suggests that near real-time deployment is feasible for compact models, while larger B3 models achieve $7.6$-$6.5$ FPS and may require optimization or specialized hardware for real-time applications. 
The throughput reduction of $15$-$17\%$ represents a reasonable trade-off given the significant improvements in temporal consistency, especially for safety-critical automated driving applications where prediction stability is paramount. 
Notably, the overhead scales consistently across different base architectures, demonstrating the general applicability of the STA mechanism without architecture-specific tuning requirements. 

\section{\uppercase{Conclusions}}
\label{Chapter:Conclusions}

This work presented a novel spatio-temporal attention (STA) mechanism that extends transformer architectures to process multi-frame video sequences for semantic segmentation. By integrating temporal reasoning directly into attention computations, STA provides a unified architectural enhancement that maintains efficiency while leveraging cross-frame correlations. Unlike optical flow-based methods or CNN-based temporal modules, STA avoids additional computational overhead and long-range dependency issues, offering a lightweight and scalable solution.

We demonstrated the effectiveness of STA across two state-of-the-art architectures (SegFormer and UMixFormer) and model scales (B0 and B3), confirming its broad applicability. Evaluation on the Cityscapes and BDD100k datasets showed consistent improvements in spatial accuracy (up to \(1.76\) pp mIoU) and temporal consistency (up to \(9.20\) pp mTC). Ablation studies further identified an optimal temporal context of \(T=3\), balancing performance gains with computational efficiency. These results highlight the suitability of STA for real-time, resource-constrained scenarios such as autonomous driving, where temporal stability is critical for safety.

While STA builds on existing attention mechanisms, its contribution lies in its simplicity, scalability, and generalizability. Overall, STA represents a practical step forward in integrating temporal reasoning into transformer-based models, making it well-suited for real-world deployment. Future work will extend STA to other video tasks, including object tracking and action recognition, and explore more efficient temporal modeling strategies for real-time applications.


\bibliographystyle{apalike}
\bibliography{references}

\end{document}